\documentclass[review]{elsarticle}

\usepackage{lineno,hyperref}
\modulolinenumbers[5]

\usepackage{subfigure}
\usepackage{graphicx}
\usepackage{amsmath,amssymb,amsthm,bm}
\usepackage{mathrsfs}
\usepackage{amsfonts}
\usepackage[none]{hyphenat}
\usepackage{multirow,soul,ulem}
\usepackage{color}
\usepackage{booktabs}

\makeatletter
\def\ps@pprintTitle{%
   \let\@oddhead\@empty
   \let\@evenhead\@empty
   \let\@oddfoot\@empty
   \let\@evenfoot\@oddfoot
}
\makeatother
\journal{Journal of Pattern Recognition Letters}

\begin{document}

\begin{frontmatter}

\title{Hallucinating very low-resolution and obscured face images}
\author[add.]{Lianping Yang\fnref{fn1}}

\fntext[fn1]{yanglp@mail.neu.edu.cn}

\author[add.]{Bin Shao}
\author[add.]{Ting Sun}
\author[addb.]{Song Ding}
\author[add.]{Xiangde Zhang\corref{mycorrespondingauthor}}
\cortext[mycorrespondingauthor]{Corresponding author:zhangxiangde@mail.neu.edu.cn}

\address[add.]{College of Sciences, Northeastern University, Shenyang 110819, China}
\address[addb.]{Beijing EyeCool Technology Co., Ltd. , Beijing 100085£¬ China}

%
%

\begin{abstract}
  Most of the face hallucination methods are designed for complete inputs. They will not work well if the inputs are very tiny or contaminated by large occlusion. Inspired by this fact, we propose an obscured face hallucination network(OFHNe-
  t). The OFHNet consists of four parts: an inpainting network, an upsampling network, a discriminative network, and a fixed facial landmark detection network. The inpainting network restores the low-resolution(LR) obscured face images. The following upsampling network is to upsample the output of inpainting network. In order to ensure the generated high-resolution(HR) face images more photo-realistic, we utilize the discriminative network and the facial landmark detection network to better the result of upsampling network. In addition, we present a semantic structure loss, which makes the generated HR face images more pleasing. Extensive experiments show that our framework can restore the appealing HR face images from $1/4$ missing area LR face images with a challenging scaling factor of $8\times$.
\end{abstract}

\begin{keyword}
\texttt Face hallucination; Obscured face images; Generative Adversarial Network (GAN)

\end{keyword}

\end{frontmatter}

\section{Introduction}
Face images are widely used in many visual applications, such as face recogni-
tion \cite{Schroff_2015_CVPR}, face alignment \cite{bulat2017far}, face parsing \cite{liu2015multi}, face expression analysis \cite{Pantic2004Facial} and so on. However, when face images are very tiny or even have a large occlusion region, many computer vision tasks will fail. It is a key technology to ease these problems by using face super-resolution(SR). Face SR, also called face hallucination(FH) can generate HR face images from LR face images. Based on deep learning, many face hallucination algorithms have been represented \cite{Huang,paper01,paper02,paper03,paper04} and achieved state-of-the-art in the last two decades.

\begin{figure}
\begin{center}
\includegraphics [width=\textwidth] {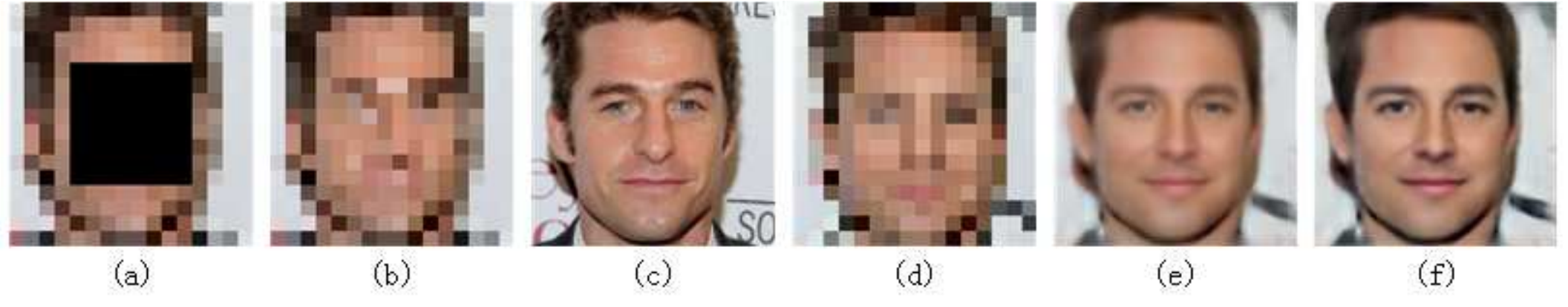}
\end{center}
   \caption{Qualitative illustration of the task. (a) $16\times16$ missing region input image. (b) $16\times16$ original LR image. (c) $128\times128$ original HR image. (d) The completed LR image by inpainting network(IN). (e) OFHNet use L2 loss only. (f) OFHNet. }
\label{introduction}
\end{figure}

Many FH methods require LR images are not too small and complete. These FH methods have achieved great success when the scaling factors are limited up to $2\thicksim 4\times$. $n\times$ scaling factor implies that we need reconstruct $n\times n$ pixels from a single pixel. One of the challenges of face hallucination is how to obtain a good HR face image when the scaling factor is $8\times$ or more. Nowadays, when we need $8\times$ or more, many methods become fragile because only little information is available. It is an extremely ill-posed and underdetermined problem.

Besides the high scaling factor, the performance of most existing FH methods will degrade dramatically when the input face image has a large occlusion region. For example, the URDGN \cite{paper01} can hallucinate aligned and complete LR frontal face images, but it cannot generate complete face images if the LR inputs with square occlusion.

To hallucinate LR inputs with occlusion, we utilize shallow encoder-decoder inpainting network to restore LR face images firstly. As shown in Fig. \ref{introduction}(d), the inpainting network can roughly repair missing region semantically. The encoder-decoder network framework has achieved great success in the field of image completion \cite{Pathak_2016_CVPR,GFC-CVPR-2017,yu2018generative}. Then we use the upsampling network to super-resolve these complete LR face images with the challenging scaling factor $8\times$. As can be seen in the Fig. \ref{introduction}(e), the generated HR face image is pleasing and recognizable. We find the inpainting network is very important in hallucinating LR face images with large occlusion. In fact, by using the LR inpainting network, the missing area can be more connected with the known area. And this strategy will reduce the difficulty of super-resolving LR occlusion face images. During phases of repairing LR face images and generating HR face images, L2 loss is used respectively. Because MSE-based deep models also generate smooth HR face images, we add not only an adversarial loss but also a semantic structure loss in upsampling network to generate more photo-realistic HR complete face images.

The adversarial loss is both widely used in general images SR \cite{Ledig_2017_CVPR} and in FH \cite{GFC-CVPR-2017,CT-FSRNet-2018,paper01,TDAE}. However, prior knowledge of face structure is also very important. Facial landmark is a very import face structure representation and hence is a kind of good prior knowledge. For example, Yu et al. \cite{Yu_2018_ECCV} use facial component heatmaps to generate more pleasing HR face images. Chen et al. \cite{CT-FSRNet-2018} use facial landmark heatmaps and parsing maps to hallucinate unaligned LR face images. Li et al. \cite{GFC-CVPR-2017} use a fixed pre-trained face parsing network to predict the face parsing labels of the generated HR face images and the corresponding ground-truth face images, and then use cross-entropy loss to ensure their face structures more similar.

Being different from \cite{GFC-CVPR-2017}, we employ facial landmark priors as our networks semantic regularization. The difference between \cite{CT-FSRNet-2018,Yu_2018_ECCV} is that we use the prior of human face as a loss function(called semantic structure loss) instead of middle features. We incorporate the adversarial loss and the semantic structure loss into the upsampling network finally. It is the final obscured face hallucination network(OFHNet). As shown in Fig. \ref{introduction}(f), the image is more photo-realistic than the image produced by adding the L2 loss only.

In summary, the contributions of this paper include: 1. We design a novel framework based on deep learning to hallucinate LR and occluded face images ($16\times16$ pixels) by a challenging scaling factor of $8\times$. 2. By using semantic structure loss, our OFHNet can generate HR complete and pleasing face images. We not only exploit local intensity information in FH but also take face structure into account. 3. Different from most FH frameworks using directly decoder networks, we present a new encoder-decoder-decoder framework and hallucinate LR incomplete face images by two steps. The first step is to complete LR obscured face inputs by LR inpainting network, and the second step is to generate HR complete and photo-realistic outputs by the upsampling network.

\section{Related work}
\subsection{Facial Structure Priors}
Although \cite{paper01, TDAE} can do FH of LR face images well, they ignore the facial structure priors. Chen et al. \cite{CT-FSRNet-2018} use a prior information estimation network to estimates landmark heatmaps parsing maps. And then they concatenate these facial structure priors with another image features to rebuild the HR face images finally. Yu et al. \cite{Yu_2018_ECCV} train a multi-task convolutional neural network to predict facial component heatmaps and output HR face images. In addition to using face prior information as middle features in networks, face prior information can also be regarded as semantic structure constraints. Li et al. \cite{GFC-CVPR-2017} present a semantic parsing loss to ensure the generated HR images are more like corresponding ground-truth images.
\subsection{Face Hallucination}
 Face hallucination is a class-specific super-resolution method. There are mainly two directions for face super-resolution reconstruction, one is the method of non-deep learning, and the other is the method of deep learning.
 In the field of non-deep learning hallucination, Zhang et al. \cite{paper02} developed a two-step statistical modeling approach that integrated a global parametric model and a local non-parametric model. The first step is to capture the common structural properties and the second step is to get more realistic details. Baker et al. \cite{paper03} proposed an algorithm to learn the gradient prior of inputs, which can be incorporated into a resolution enhancement algorithm. In \cite{Yang2013Structured}, considering that each face image is represented in terms of facial components facial components, contours, and smooth regions, they proposed a structured face hallucination(SF-
 H) method that maintains the image structure by matching gradients in the rebuild high-resolution output. However, this method relies on the accuracy of facial key points position. When the size of the input image is smaller, the performance of SFH will dramatically reduce.

 In the field of deep learning hallucination, Yu and Porikli \cite{paper01} proposed a discriminative generative network to super-resolve $8\times$ face images. The generator reconstructs a smooth HR image by L2 loss and the discriminator uses an adversarial loss to make the hallucinated HR image more similar to real one. Tuzel et al. \cite{Tuzel2016Global} developed another deep learning framework which learns global and local constraints of the HR faces jointly. \cite{Cao2017Attention}, developed by \cite{Tuzel2016Global}, used deep reinforcement learning to get the more sharp result. However, all these methods need complete faces to guarantee good results.
\subsection{Auto-encoder and Image completion}
 Stacked Denoising Autoencoders \cite{Vincent2010Stacked} can learn more useful compact represen-
 tations from encoder than traditional auto-encoders. Denoising auto-encoders corrupt the image and force the network to undo the damage. By this way, the network can learn more robust feature representation which promotes reconstruc-
 tion tasks. Image completion is a work that has attracted much attention in the field of computer vision \cite{paper05,Pathak_2016_CVPR}. \cite{paper05}combined sparse coding with stack denoising auto-encoder(DA) to solve the problem of blind image inpainting. When the size of images is small like $16\times16$, less DAs have good results. \cite{Pathak_2016_CVPR} shared a similar encoder-decoder architecture which has an adversarial loss to complete the input with large missing region. Like encoder-decoder architecture of auto-encoders, we use a shallow LR inpainting network to complete inputs roughly.

\begin{figure}
\begin{center}
\includegraphics [width=\textwidth] {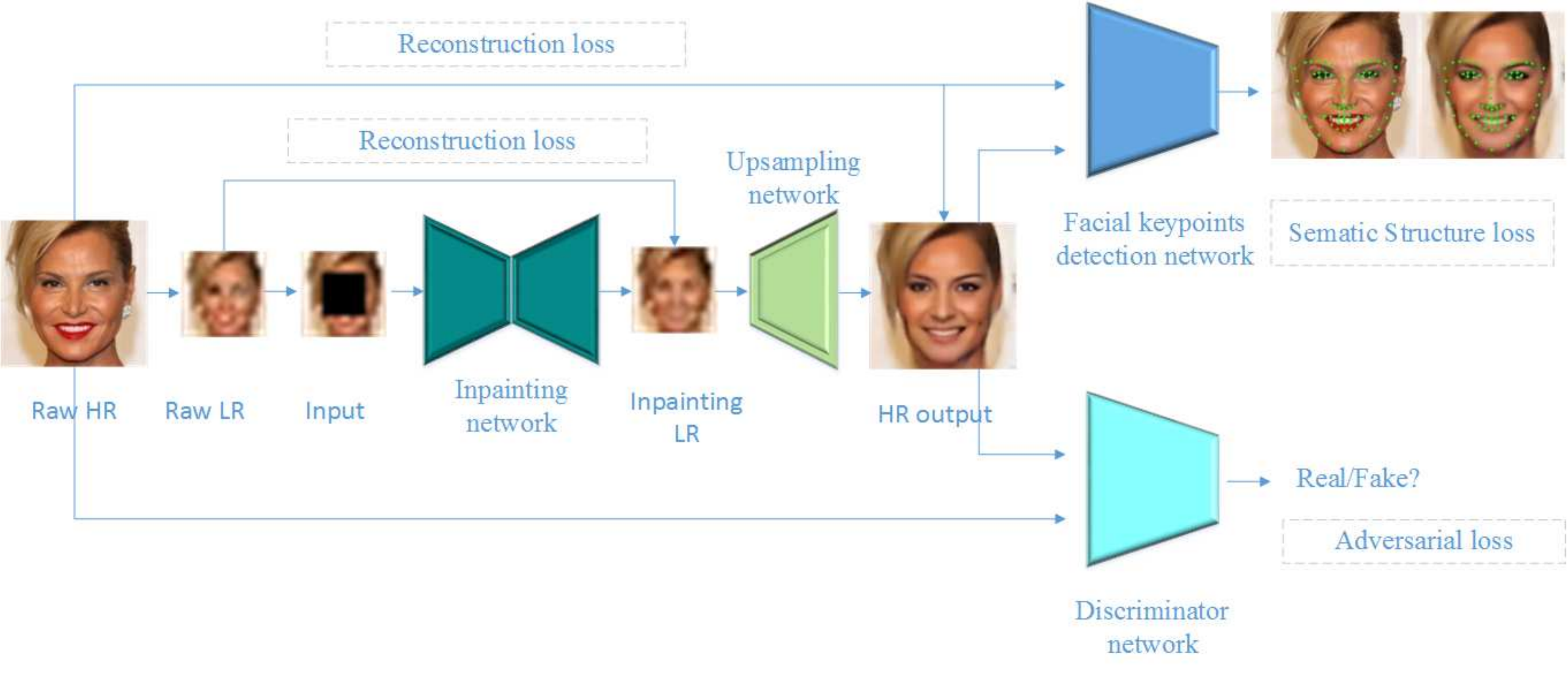}
\end{center}
   \caption{Training workflow of OFHNet. The OFHNet consists of four parts: an inpainting network, an upsampling network, a discriminator network, and a fixed facial keypoints detection network. The inpainting network restores the input firstly in order to alleviate upsampling network's burden of face hallucination. We replace pixels in the non-mask region of the inpainting LR with original pixels. Upsampling network takes the inpainting LR as input and exports the HR and complete face image. The discriminator distinguishs real and generated complete HR output globally. The fixed facial keypoints detection is a pre-trained model, and is to further improve the HR output more photo-realistic.
}
\label{pipline}
\end{figure}

\begin{figure}
  \centering
  \includegraphics[width=0.6\textwidth]{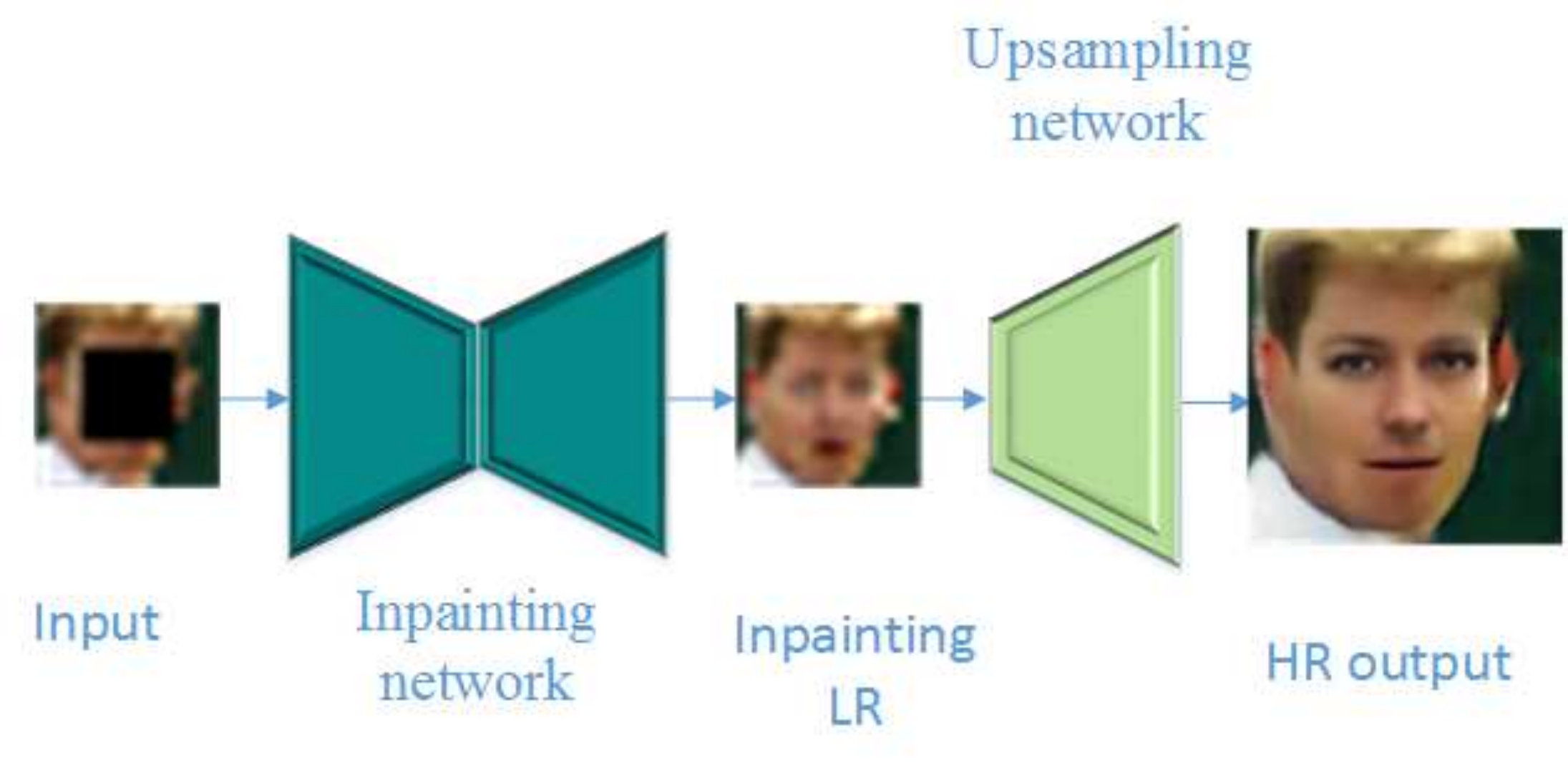}
  \caption{Testing workflow of our OFHNet.}\label{testWorkflow}
\end{figure}

\section{Proposed Algorithm}
Our framework has four components, an LR image inpainting network, an upsampling network, a discriminative network, and a facial landmark detection network. Our goal is to hallucinate aligned LR face images with large occlusion. As shown in Fig. \ref{pipline}, we first use LR inpainting network to complete LR obscured face inputs. Secondly, we use the upsampling network to hallucinate the comple-
te LR face images. As shown in Fig. \ref{testWorkflow}, we use two steps to train our OFHNet, but it can end-to-end to generate complete HR face images in testing.

\begin{figure*}
\begin{center}
\includegraphics [width=\textwidth] {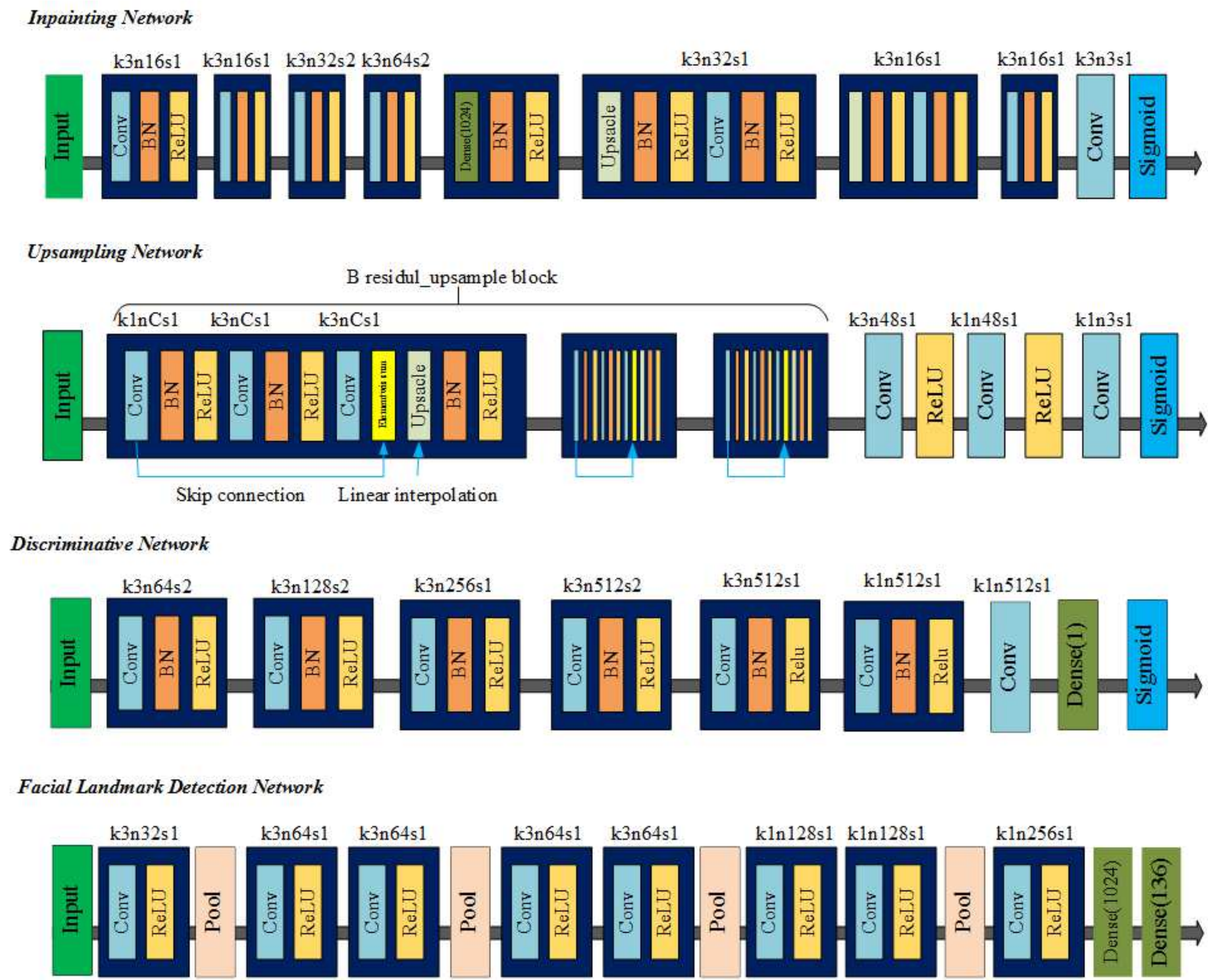}
\end{center}
   \caption{Architectures of our OFHNet's four components.}
\label{four_networks}
\end{figure*}
\subsection{Network Architecture}
\paragraph{Inpainting Network} As shown in Fig. \ref{four_networks}, we design a shallow encoder-decoder network to complete LR inputs. We stack convolutional, batch-normalization and ReLU layers(Conv+ReLU+BN) as a convolutional block. The encoder of the inpainting network consists of four convolutional blocks, the first two convolutional block with stride 1, the next two convolutional block with stride 2. Symbol (k3n16s1) defines the convolutional layer of a block with 3 kernel size (k), 16 numbers of feature maps (n) and 1 stride(s). After encoder, we obtain useful and compact bottle representations(the dimension is 1024). Our decoder consists of two similar upsample modules but the numbers of maps are different. The upsample modules stack nearest-neighbor interpolation, batch-normali-
zation, ReLU, convolutional, batch-normalization and ReLU layers(NNI+BN+
ReLU+Conv+BN+ReLU). We utilize nearest-neighbor interpolation instead of deconv \cite{Radford2015Unsupervised} as our upsample operation for the reason that it can reduce checkerboard pattern of artifacts \cite{Odena2016Dev}. Finally, two convolutional layers and a sigmoid layer are used to produce LR complete output.

\paragraph{Upsampling Network} Upsampling network has three residual-upsample blocks, each increases 2x spatial resolution and total are 8x. Residual-upsample blocks employ skip connect to reduce gradient dispersion \cite{paper11}. Each residual-upsample block consists of convolutional, batch-normalization, ReLU, convolutional, batch
-normalization, ReLU, convolutional, elementwise sum, nearest-neighbor interp-
olation, batch-normalization and ReLU layers(Conv+BN+ReLU+Conv+BN+
ReLU+Conv+Ele+NNI+BN+ReLU). All convolutional layers of every residual-
upsample blocks have the same number of feature maps. Symbol (k1nCs1) indicates this block have C number feature maps. After three convolutional layers, we use sigmoid layer as the output layer.

\paragraph{Discriminative Model} We use the pair of fake face image and its real image as the input of discriminator to calculate the probability that the image is judged to be true. The discriminator output value is a scale between 0 and 1. We use six convolutional layers followed by a batch-normalization layer and ReLU layer, finally connected to a convolutional, a densenet and a sigmoid layer.

\begin{figure}
  \centering
  \includegraphics[width=0.5\textwidth]{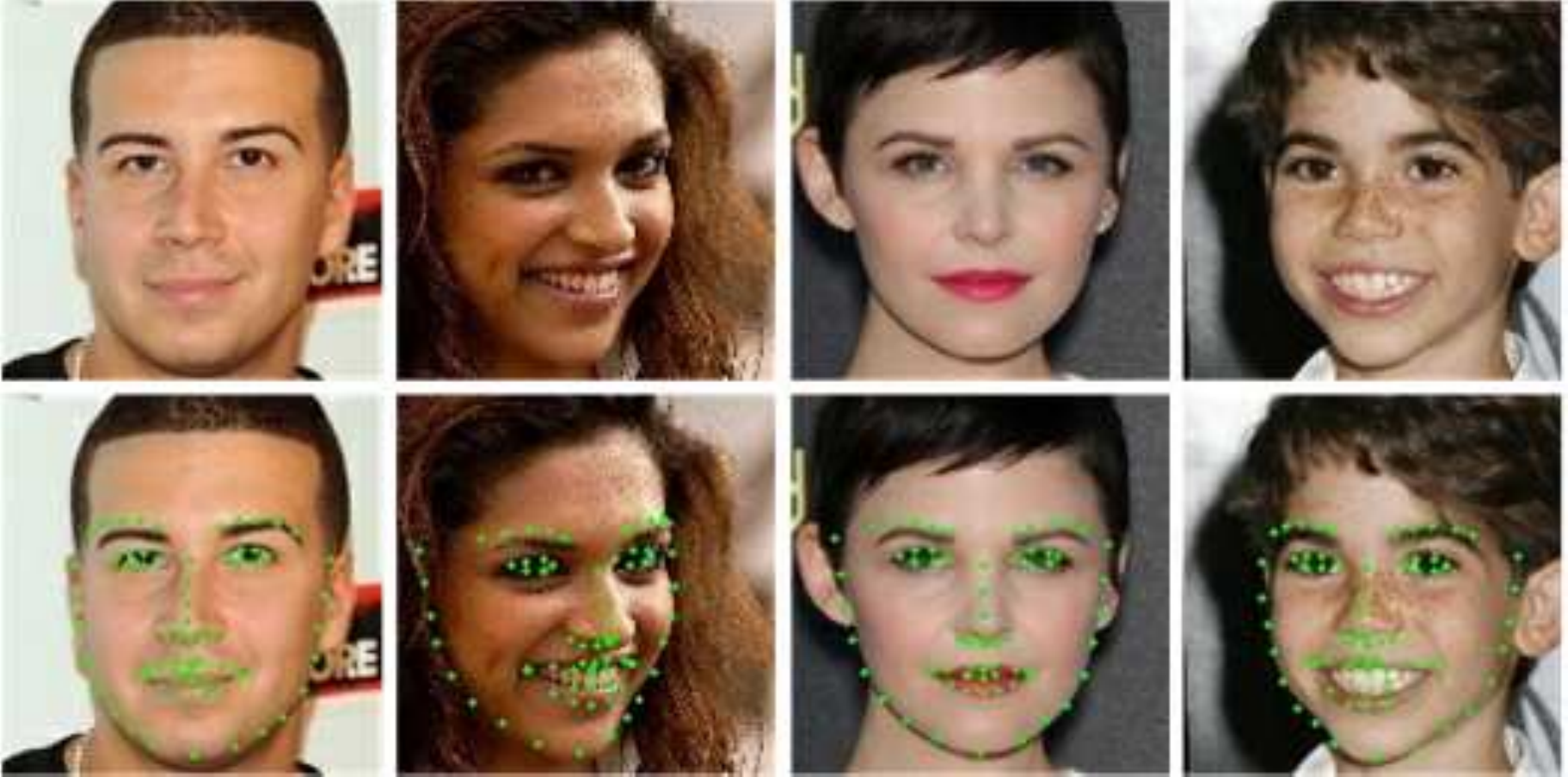}\\
  \caption{Examples of landmark detection result on CelebA by pre-trained facial landmark detection network. The first row is the original HR image and the second row is the result.}\label{landmark}
\end{figure}

\paragraph{Facial Landmark Detection Network} We adopt the first level convolutional neural network(CNN) of \cite{Fan2016Approaching} as our facial landmark detection network. As shown in Fig. \ref{landmark}, there are some results by the pre-trained landmark detection network.

\subsection{Loss Function}
Given the training set $\{ {l_i},{\hat l_i},{h_i}\},i=1,2....N$, where $l_i$ represents the i-th LR input face image with large missing region, $\hat l_i$ represents the i-th LR label face image and $h_i$ represent the corresponding i-th HR clean face image.

\paragraph{Reconstruction Loss}  To make the completed LR face images and the generated HR complete face image as close as possible to the corresponding ground truth face images, respectively, we adopt two L2 losses to enforce their similarity. For the inpainting network, it is as follows:

\begin{equation}
   {\ell _{{R_1}}}\left( w \right) = \frac{1}{N}\sum\limits_{i = 1}^N {\left( {{{l'}_i} - {{\hat l}_i}} \right)}^2  = \frac{1}{N}{\sum\limits_{i = 1}^N {\left( {{I_w}\left( {{l_i}} \right) - {{\hat l}_i}} \right)} ^2}     \label{eq1}
\end{equation}

where ${l'_i}$ and ${I_w}\left( {{l_i}} \right)$ both represent the completed LR faces by inpainting network, $N$ is the numbers of training image pairs$\{ {l_i},{\hat l_i},{h_i}\}$, w is the parameters of inpainting network. For the L2 loss of upsampling network, it is as follows:

\begin{equation}
  {\ell _{{R_2}}}\left( t \right) = \frac{1}{N}{\sum\limits_{i = 1}^N {\left( {{h_i}^\prime  - {h_i}} \right)} ^2} = \frac{1}{N}{\sum\limits_{i = 1}^N {\left( {{U_t}\left( {{l_i}^\prime } \right) - {h_i}} \right)} ^2}    \label{eq2}
\end{equation}

where ${h_i}^\prime$ and ${U_t}\left( {{l_i}^\prime } \right)$ both represent the completed HR faces by upsampling network and t is the parameters of upsampling network.

\paragraph{Adversarial Loss} The L2 loss will generate over-smoothed SR outputs \cite{CT-FSRNet-2018}. To solve this problem, we add adversarial loss to train upsampling network. The purpose of adversarial loss is to make the generated data distribution as close as possible to the real data distribution. In our case, we need to make the generated face images and real face images as similar as possible. It has two components: the discriminator D and the generator G. We use the upsampling network as our generator. The objective function for the discriminator D is formulated as:

\begin{equation}
  \ell _{adv}^D\left( d \right) = \frac{1}{N}\sum\limits_{i = 1}^N {\left[ {\log {D_d}\left( {{h_i}} \right) + \log \left( {1 - {D_d}\left( {{{h'}_i}} \right)} \right)} \right]}  \label{eq3}
\end{equation}

where d is the parameters of discriminator network. The discriminator is to discriminate the input face image correctly, but the generator is to fool the discriminator. The objective function for the generator is formulated as:

\begin{equation}
  \ell _{adv}^U\left( t \right) =  - \frac{1}{N}\sum\limits_{i = 1}^N {\log {D_d}} \left( {{U_t}\left( {{l_i}} \right)} \right) \label{eq4}
\end{equation}

where t is the prameters of upsampling network.

\paragraph{Semantic Structure Loss} In order to further enhance the structural similarity between generated face images and corresponding real face images, the informati-
on of facial landmarks is utilized. The first level CNN of \cite{Fan2016Approaching} is employed as our facial landmark detection network. The facial landmark detection network can produce 68 landmark points and uses Euclidean distance as its objective function. To train the upsampling network by semantic structure loss, we pre-train the landmark detection network and then fix the parameters. Semantic structure loss is formulated as:
\begin{equation}\label{eq5}
  {\ell _S}\left( t \right) = \frac{1}{N}{\sum\limits_{i = 1}^N {\left( {A\left( {{h_i}^\prime } \right) - A\left( {{h_i}} \right)} \right)} ^2}
\end{equation}

where A is the fixed facial landmark detection network. Note that, we only use the last layer of facial landmark detection network.
\subsection{Training Details}
 In order to train our network effectively, we use curriculum strategy \cite{Bengio2009Curriculum} by dividing the training process into two steps. Firstly, in training our LR inpainting network, we minimize Eqn. \ref{eq1} to update the parameters w. Secondly, in training our upsampling network, multiple losses, $i.e.$, ${\ell _{{R_2}}}\left( t \right)$, $\ell _{adv}^U\left( t \right)$, ${\ell _S}\left( t \right)$, are incorporated into a total loss function ${\ell _\Gamma }$. So the final objective function of the upsampling network is:

\begin{equation}\label{eq_u}
  {\ell _\Gamma } = {\ell _{{R_1}}} + \alpha \ell _{adv}^U + \beta {\ell _S}
\end{equation}

where $\alpha$, $\beta$ are the trade-off weights. Same as \cite{Yu_2018_ECCV}, our discriminator network and upsampling network are trained in an alternating fashion, simultaneously. We also maximize the loss Eqn. \ref{eq3} and minimize the loss Eqn. \ref{eq_u} to update parameters $d$ and $t$, respectively.

\section{Experiments}
\subsection{Datasets}
Large-scale CelebFaces Attributes (CelebA) dataset \cite{liu2015faceattributes} is employed in our experiments. It consists of more than 200K celebrity images with large pose variations and background clutter. Like \cite{TDAE}, we choose 28000 images randomly as training dataset and 2000 images as testing dataset. Note that, training dataset and testing dataset are not duplicated. Firstly, we use five landmark locations to align faces. Secondly, we crop the images by $128*128$ and use cubic interpolation method downsamples the images. We occlude the LR image randomly, and the occlusion area is larger than $1/16$ and less than $1/4$ of the LR image. Since CelebA only has a ground truth of 5 landmarks, we utilize a highly accurate facial landmark localization algorithm \cite{bulat2017far} to estimate the 68 landmarks as the ground truth to pre-train the facial landmark detection network.

 \begin{figure*}[ht]
\centering
\includegraphics[width=\textwidth]{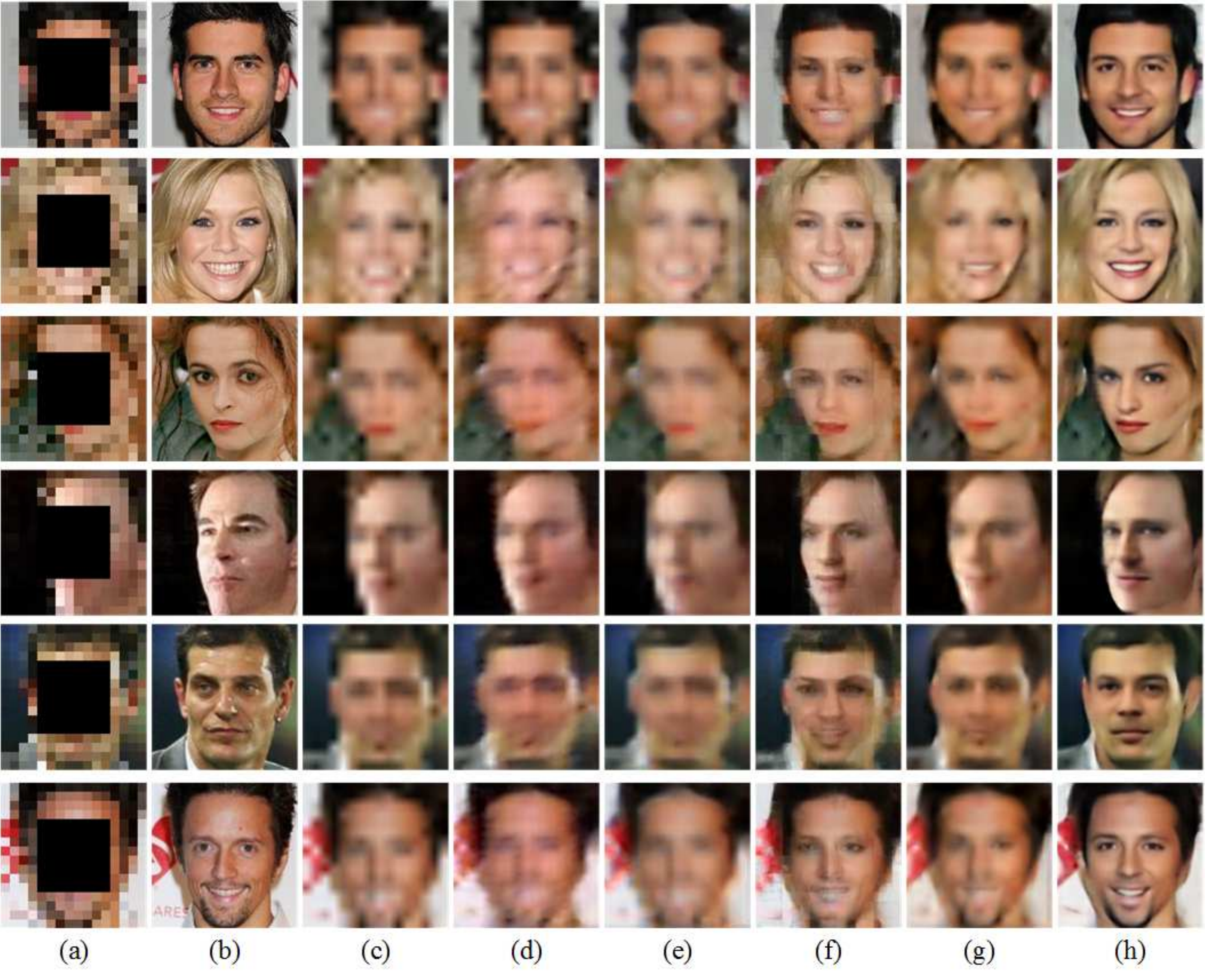}
\caption{Comparison with state-of-the-art methods. (a)LR missing region input image. (b)Original HR images. (c)IN+bicubic. (d)IN+FSRCNN (e)IN+VDSR. (f)IN+Ma. (g)IN+URDGN. (h)our OFHNet.}
\label{fig:pathdemo}
\end{figure*}

\subsection{Evaluation protocols and comparisons}
 We not only compare with several face hallucination approaches, but also compare with some super-resolution methods. For face hallucination, we chose the method in \cite{Ma2010Hallucinating} which uses the optimal weights of the training image position-
 patches to hallucinate face. This method needs complete aligned face images. Based on the sample learning method, Yu et al. \cite{paper01} ultra-resolve face images by discriminative generative networks. As for general super-resolution methods, we chose \cite{VDSR} and \cite{FSRCNN}. For fair comparison, we use our inpainting network(IN) to complete the input for all methods.

 \subsection{Qualitative and quantitative comparisons}
  We use our IN to complete the missing regions of input, then use bicubic interpolation to hallucinate faces. The results are shown in Fig. \ref{fig:pathdemo}(c). These images are very blurry and without obvious outline.

FSRCNN \cite{FSRCNN} is a fast SR method but do not support upscaling factor 8x. We set the deconvolution layer of FSRCNN with stride 8 to support upscaling factor 8x. Because the size of input is very small, it is difficult for FSRCNN to hallucinate the face image well. As shown in Fig. \ref{fig:pathdemo}(d), the image looks vague and cannot be recognized the expression.

  Kim et al. present a very deep network \cite{VDSR} to reconstruct general LR image. The deep network is used to learn a residual image. In Fig. \ref{fig:pathdemo}(e), the result is very smooth.

\begin{table}
\begin{center}
\caption{Comparison between our method with others in terms of PSNR and SSIM.}

\begin{tabular}{lll}
\hline\noalign{\smallskip}
Method & PSNR & SSIM  \\
\noalign{\smallskip}\hline\noalign{\smallskip}
  IN+bic & 17.53 & 0.44 \\
  IN+Ma &17.36 &0.44\\
  IN+VDSR &19.60 & 0.53\\
  IN+FSRCNN & 19.78 & 0.54 \\
  IN+URDGN & 20.00 & 0.57 \\
  OFHNet & 20.44 & 0.63 \\
\noalign{\smallskip}\hline
\end{tabular}
\label{lab_1}       
\end{center}
\end{table}

  Ma et al. \cite{Ma2010Hallucinating} develop a local model using mapping function between LR position-patch and HR position-patch. As is shown in Fig. \ref{fig:pathdemo}(f) , after we use IN to restore the input first, the result shows it is not able to maintain the detailed local features.

  As illustrated in Fig. \ref{fig:pathdemo}(g), based on IN, URDGN \cite{paper01} can hallucinate face images with missing region. But URDGN still generate HR images without high-frequency facial details.

  In Fig. \ref{fig:pathdemo}(h), the OFHNet super-resolves obscured LR face images and in contribute to more pleasing and realistic HR face images than other methods.

  The Tab. \ref{lab_1}. shows the comparison of our method with other state-of-the-art methods about PSNR and SSIM. The bottom line is the result of OFHNet over URDGN. The PSNR value is 0.44 dB greater than URDGN and the SSIM imporves 10.5\%.

\begin{figure}
  \centering
  \includegraphics[width=0.5\textwidth]{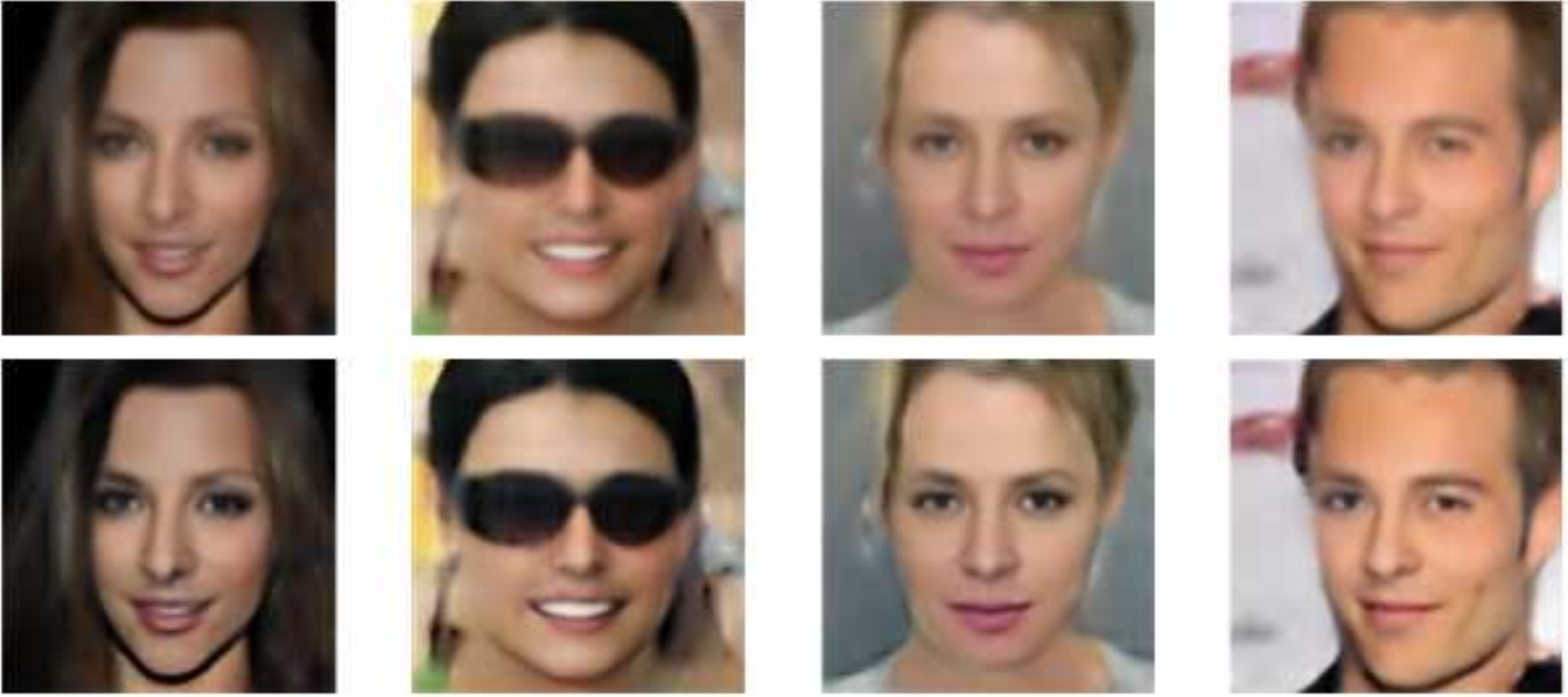}
  \caption{Visual effectiveness improved by semantic structure loss. The first row only use L2 loss. The second row use L2 loss + semantic structure loss without adversarial loss.}\label{SSL:effect}
\end{figure}

\subsection{Analysis and Discussion}
\paragraph{Visual Effectiveness of Semantic Structure Loss} In order to show semantic loss can improve the visual effect, we use L2 loss and L2 loss + semantic structure loss to train upsampling network respectively, without considering adversarial loss. As is shown in Fig. \ref{SSL:effect}, the visual effect is improved.

\begin{table}[b]
\begin{center}
\caption{Quantitative comparisons of OFHNet and URDGN on alignment (NRMSE)/parsing (IoU). Note that, the smaller the NRMSE is better and the bigger the IoU is better.}

\begin{tabular}{lll}
\hline\noalign{\smallskip}
  Method & NRMSE & IoU  \\
\noalign{\smallskip}\hline\noalign{\smallskip}
  IN+URDGN & 13.61 & 0.40 \\
  OFHNet & 5.33 & 0.60 \\
\noalign{\smallskip}\hline
\end{tabular}
\label{tab_2}       
\end{center}
\end{table}

\paragraph{The Quality of Geometry in Generated Face images} Although PSNR and SSIM are widely used, they can not focus on the photometric quality of the generated image very well \cite{CT-FSRNet-2018}. So we employ face alignment and parsing as new evaluation metrics like \cite{CT-FSRNet-2018} to compare our OFHNet with URDGN. We adopt the algorithms in \cite{bulat2017far} and \cite{GFC-CVPR-2017} to obtain landmark points and predict the parsing maps for the recovered images of OFHNet and URDGN. As shown in Tab. \ref{tab_2}, the quality of geometry in generated images by OFHNet is better than URDGN.

\begin{figure}
  \centering
  \includegraphics[width=\textwidth]{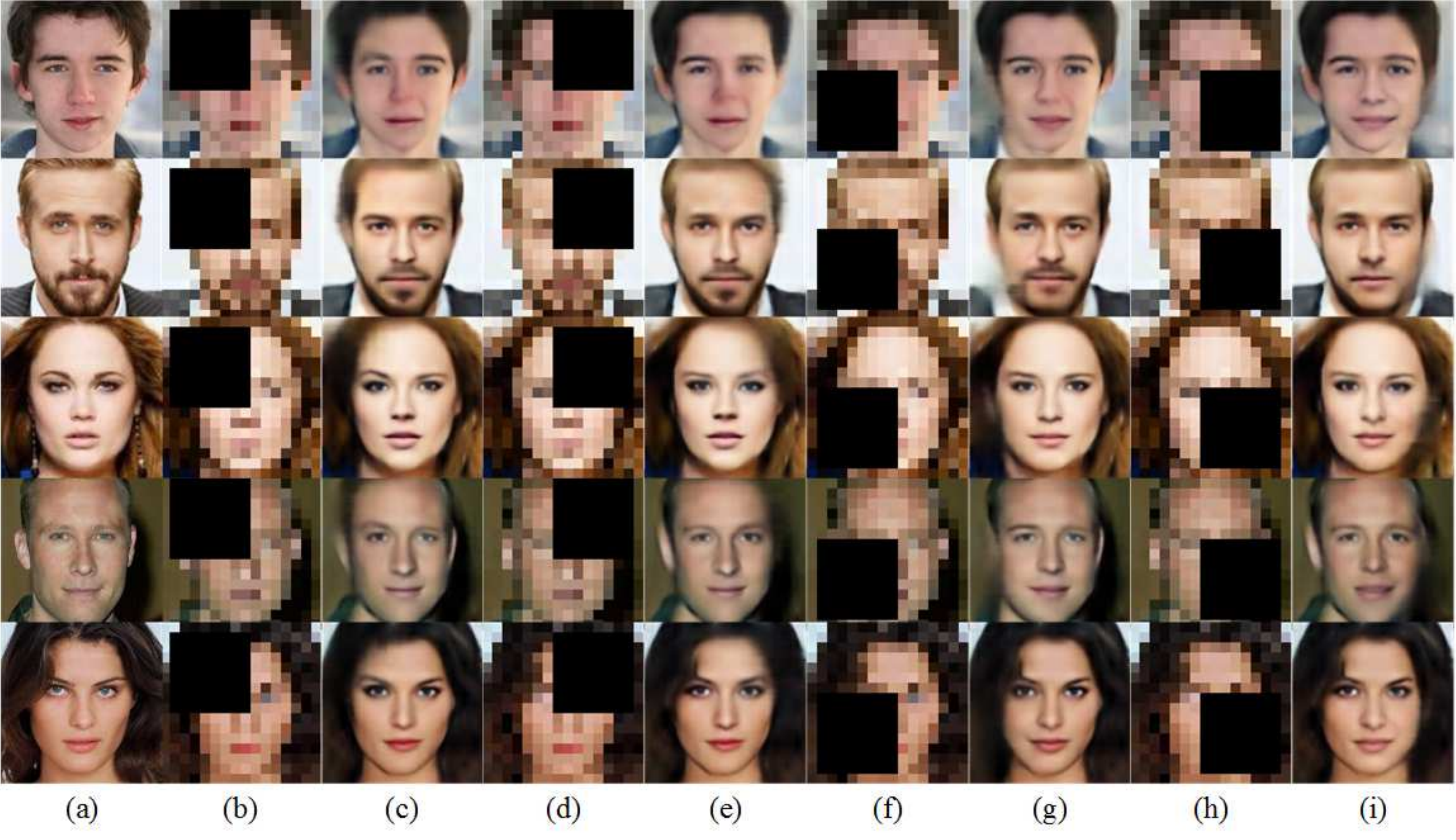}\\
  \caption{Visual quality of generated results under random occlusion. (a) HR original image. (b) LR input with top-left occlusion (c) recovered HR image of LR input with top-left occlusion. (d)LR input with top-right occlusion. (e) recovered HR image of LR input with top-right occlusion. (f)LR input with down-left occlusion. (g) recovered HR image of LR input with down-left occlusion. (h)LR input with down-right occlusion. (i) recovered HR image of LR input with down-right occlusion.}\label{random_o}
\end{figure}

\paragraph{Random Occlusion} As shown in Fig.\ref{random_o}, we set up four random occlusions: top-left, top-right, bottom-left, and bottom-right. The occlusion size is $1/4$ of LR obscured input. We can see that our OFHNet repair these random occlusion areas well and generates pleasing results.

\begin{figure}
  \centering
  \includegraphics[width=\textwidth]{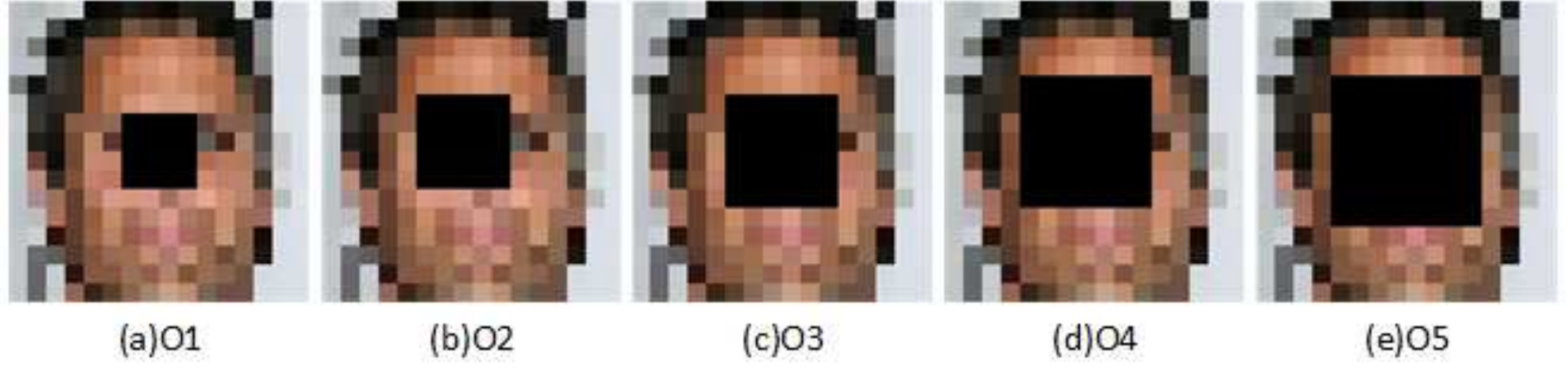}
  \caption{Different sizes of face occlusion in the central area O1-O5. From left to right: $4\times4$(occlusion size), $5\times5$, $6\times6$, $7\times7$, $8\times8$.}\label{fig6}
\end{figure}

\begin{figure}[ht]
  \centering
  \includegraphics[width=0.6\textwidth]{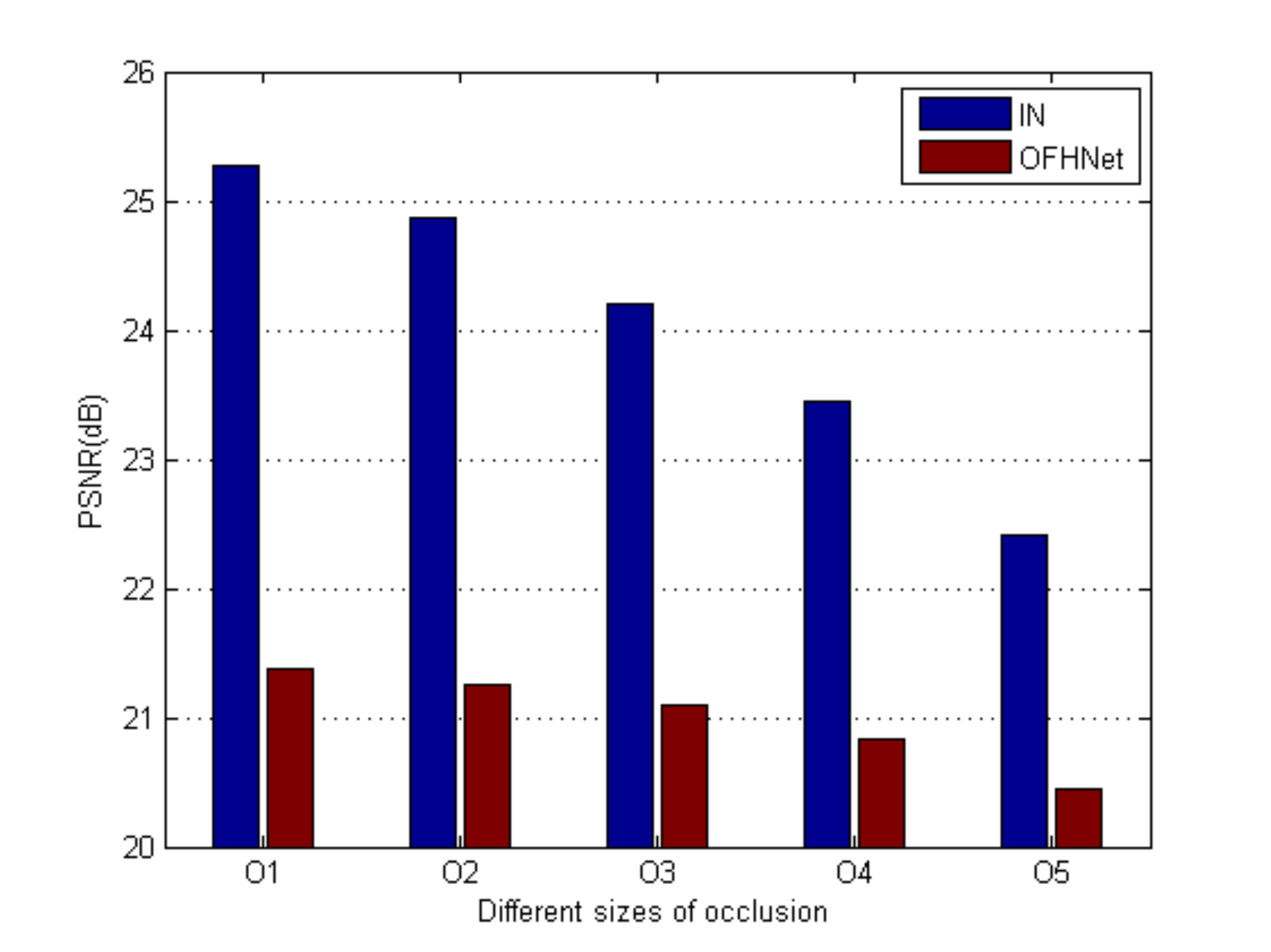}
  \caption{Ablation study on effects of different occlusion. The outputs of IN compare with original LR face images. The outputs of OFHNet compare with original HR face images.}\label{fig_7}
\end{figure}

\paragraph{Effects of Different Sizes of Occlusion} For different sizes of occlusion in the central area, we analyze the inpainting network and OFHNet. As shown in Fig.\ref{fig6}, we set five different sizes of face occlusion in the central area. As the increasing of occlusion size, the difficulty of reconstruction is increasing(Fig.\ref{fig_7}). The PSNR values are monotonically decreasing.

\section{Conclusion}
We have presented a new and effective obscured face hallucination network(
OFHNet) to super-resolve LR face images with $1/4$ occlusion by a challenge factor of $\times8$. Our OFHNet adopts an encoder-decoder-decoder framework, which can solve the problem of FH with occlusion effectively. We not only explore the image intensity correspondences between LR and HR faces, but also consider their semantic structure regularization. We find semantic structure loss can greatly enhance the visual effects of generated HR face images. In the future, we will strengthen this semantic structure regularization to improve the image quality.

\section{Conflict of Interests}
The authors declare that there is no conflict of interests regarding the public-
ation of this paper.

\section{Acknowledgment}
This work is supported by the Fundamental Research Funds for the Central Universities (Grant No. N160504007 ) and supported by the National Natural Science Foundation of China (Grant No.31301086).


\end{document}